\documentclass[runningheads]{llncs}
\usepackage{eccv}
\usepackage{eccvabbrv}
\usepackage{graphicx}
\usepackage{booktabs}
\usepackage{wrapfig}
\usepackage[ruled]{algorithm2e}
\usepackage[sort, numbers]{natbib}
\usepackage{array}
\usepackage{fancyhdr}
\usepackage{float}
\usepackage[pagebackref]{hyperref}
\usepackage{appendix}
\usepackage{orcidlink}

\usepackage{url}
\usepackage{color, colortbl}
\usepackage[font=scriptsize]{caption}
\usepackage[font=scriptsize]{subcaption}
\newcolumntype{P}[1]{>{\centering\arraybackslash}p{#1}}
\definecolor{LightCyan}{rgb}{0.9,1,1}
\definecolor{LightOrange}{rgb}{1,0.87,0.87}

\usepackage{graphicx}
\graphicspath{ {./figures/} }

\usepackage{amsmath,amsfonts,bm}

\def\eqref#1{equation~\ref{#1}}

\def\1{\bm{1}}

\DeclareMathAlphabet{\mathsfit}{\encodingdefault}{\sfdefault}{m}{sl}
\SetMathAlphabet{\mathsfit}{bold}{\encodingdefault}{\sfdefault}{bx}{n}

\usepackage[capitalize]{cleveref}
\crefname{section}{Sec.}{Secs.}
\crefname{section}{Section}{Sections}
\crefname{table}{Table}{Tables}
\crefname{table}{Tab.}{Tabs.}
\crefname{alg}{Algorithm}{Algorithms}

\title{VideoClusterNet: Self-Supervised and Adaptive Face Clustering for Videos}

\author{Devesh Walawalkar\orcidlink{0000-0001-9464-9027} \and
Pablo Garrido\orcidlink{0009-0001-8273-6737}}

\institute{Flawless AI \\
\email{\{devesh.walawalkar,pablo.garrido\}@flawlessai.com}\\
\url{https://www.flawlessai.com/}}

\authorrunning{D. Walawalkar et al.}

\begin{document}
\maketitle

\begin{abstract}

With the rise of digital media content production, the need for analyzing movies and TV series episodes to locate the main cast of characters precisely is gaining importance.Specifically, Video Face Clustering aims to group together detected video face tracks with common facial identities. This problem is very challenging due to the large range of pose, expression, appearance, and lighting variations of a given face across video frames. Generic pre-trained Face Identification (ID) models fail to adapt well to the video production domain, given its high dynamic range content and also unique cinematic style. Furthermore, traditional clustering algorithms depend on hyperparameters requiring individual tuning across datasets. In this paper, we present a novel video face clustering approach that learns to adapt a generic face ID model to new video face tracks in a fully self-supervised fashion. We also propose a parameter-free clustering algorithm that is capable of automatically adapting to the finetuned model's embedding space for any input video. Due to the lack of comprehensive movie face clustering benchmarks, we also present a first-of-kind movie dataset: MovieFaceCluster. Our dataset is handpicked by film industry professionals and contains extremely challenging face ID scenarios. Experiments show our method's effectiveness in handling difficult mainstream movie scenes on our benchmark dataset and state-of-the-art performance on traditional TV series datasets. \footnote{\label{note4} Accepted in European Conference on Computer Vision (ECCV) 2024}
\end{abstract}

\section{Introduction}\label{sec:intro}
Video Face Clustering can be defined as the task of grouping together human faces in a video among common identities. It contributes significantly to several other research domains, such as video scene captioning \cite{rohrbach2017_movie_description}, video question answering \cite{tapaswi2016_movieQA}, and video understanding \cite{vicol2018_moviegraphs}.
%
Having an understanding of the spatial location, face size, and identity of the characters that appear in specific scenes is essential for all the aforementioned tasks.
Clustering faces in a video is a challenging unsupervised problem that has garnered a lot of interest over the past few decades \cite{satoh1999_name-it,pham2009_cross-media,zhou2015_multi-cue,wu2013_constrained-clustering}. Given the rise in the creation of video production content and the subsequent need for its analysis, face clustering in the movie/TV series domain has garnered significant interest in the last couple of years \cite{sharma2019_self-supervised,tapaswi2019_video-face}. It serves as an effective editing tool for movie post-production personnel, helping them select scenes with a specific group of characters, among other benefits. We thus primarily focus on the video production content domain for evaluating our proposed method, given its closeness to real-world scenarios and use cases. 

\begin{wrapfigure}[22]{R}{0.4\textwidth} 
\includegraphics[scale=0.9,width=4.7cm]{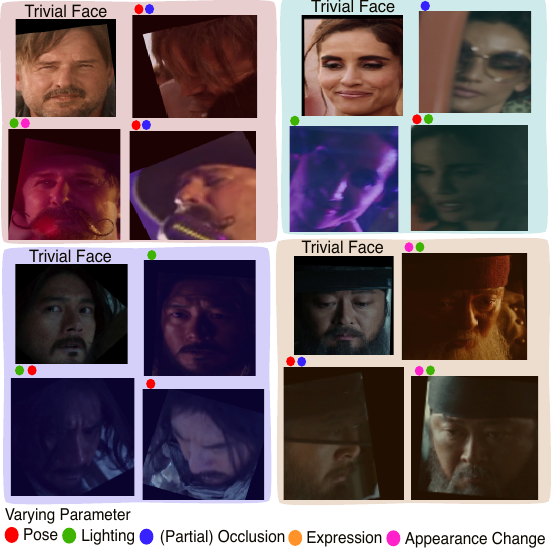}
\centering
\caption{Select hard case clusters predicted using our algorithm from within \textbf{MovieFaceCluster} dataset. Trivial face represents an easy ID sample for each cluster. The term “varying parameter” depicts the dominant image attributes that are particularly challenging for a given face crop. It is not part of the dataset annotations but is depicted for enhanced reader understanding.}
\label{fig:benchmark_hard_case_viz}
\end{wrapfigure}

The video production content domain often provides an unique set of challenges for face clustering, in terms of large variations in facial pose, lighting, expression and appearance (\cref{fig:benchmark_hard_case_viz}). 
In specific domains with high-quality standards, such as movies that possess a unique cinematic style\footnote{Particular movie features include high resolution, high dynamic range, and large facial attribute variations.},
performance of face identification (ID) models trained on generic large-scale datasets is often limited for such domains (\cref{table:ablation_face_id_model}). Furthermore, hand labeling a large cast of characters, often present in movies/TV series, can be very time-consuming and error-prone. As a result, the inherent challenges in video face clustering and difficulties in hand labeling often limit video-specific model training. In this paper, we propose an algorithm that successfully tackles these limitations. Specifically, our proposed method adapts a generic face ID model to a specific set of faces and their observed variations in a given video in a fully self-supervised fashion.     
Also, traditional deep face clustering algorithms \cite{zhou2022_clustering-survey} present certain limitations, which can be categorized into two main types. The first group \cite{defays1977_efficient_algorithm,ester1996_density_based} adopts a bottom-up approach to clustering and incorporates a pre-defined distance function to compare face embeddings, thus requiring a user-defined threshold to specify a positive match. The second group \cite{Lloyd1982_least_squares,comaniciu2002_mean_shift} follows a top-down approach and requires the number of known clusters or minimum cluster sample count as input for its initialization. Thus, both these groups have shortcomings in terms of requiring non-intuitive user-defined parameters. We present a novel agglomerative clustering algorithm that improves on these limitations. It is fully automated without the need for any user input and uses a distance metric that is optimized for the given model-learned embedding space. 

Overall, our proposed method VideoClusterNet can be divided into two main stages. The first stage involves a fully self-supervised learning (SSL) based finetuning of a generic face ID model on a given set of video faces. The finetuning task is formulated as an iterative optimization task facilitated through alternating stages of model finetuning and coarse face track matching. The SSL finetuning bootstraps itself by soft grouping together high-confidence matching tracks at regular training intervals. The second stage involves a track clustering algorithm that adopts the loss function used for SSL finetuning as a distance metric. Our clustering algorithm computes a custom matching threshold for each track and combines tracks with high-confidence matches in an iterative bottom-up style.

Given the unique in-the-wild challenges in video production domain, the academic research community lacks a standardized video dataset benchmark for real-world performance evaluation. Thus, we also present a novel video face clustering dataset, which incorporates
challenging movies hand-selected by experienced film post-production specialists.
We conduct extensive experiments of our proposed method on this dataset to validate its effectiveness for character clustering in mainstream movies. In addition, we provide results on selects benchmark datasets, showing that our method attains state-of-the-art performance.        

In summary, \textbf{we propose the following contributions:}
1) A fully self-supervised video face clustering algorithm, which progressively learns robust identity embeddings for all faces within a given video face dataset, facilitated via iterative soft matching of faces across pose, illumination, and expression variations observed in the dataset. 2) A self-supervised model finetuning approach that, unlike prior works, relies only on positive match pairs, removing any dependence on manual ground truth cluster labels or use of temporal track constraints to obtain negative match pairs. 3) A deep learning-based similarity metric for face clustering, which automatically adapts to a given model's learned embedding space. 4) A novel video face clustering algorithm that does not depend on any user-input parameters.

Additionally, we present a new comprehensive movie face clustering dataset to better evaluate video face clustering algorithms on real-world challenges.
\section{Related Work}\label{sec:related_work}

We review prior work in video-based face clustering and list out some deep learning metric and self-supervised learning based methods since they form an important component of our approach.

 \textbf{Auxiliary labels assisted Video Face Clustering:} 
Single frame-based face clustering has drawn a lot of attention in the past few decades. For a detailed survey, please refer to \cite{zhou2022_clustering-survey}. For the video domain, early work focused on using additional information available from TV series episodes/movies. Specifically, methods such as \cite{satoh1999_name-it,berg2004_names,cour2010_talking_pictures,everingham2009_taking-bite, tang2015_face-clustering,ozkan2006_graph,pham2009_cross-media,everingham2006_automatic} utilize aligned captions, transcripts, dialogues, or a combination of the above with detected faces to perform identity clustering.       

\textbf{Contextual Information based Video Face Clustering:} Following work using supplementary labels, methods such as \cite{zhang2013_unified_framework,el2010_face} leverage contextual information, e.g., clothing and surrounding scene contents, while \cite{paul2014_conditional} use aligned audio to help localize faces. Alternatively, \cite{zhou2015_multi-cue} incorporate gender information along with temporal constraints through face motion tracking. Unlike previous approaches, our method requires no explicit contextual information.

\textbf{Video Face Clustering using temporal feature aggregation, 3D convolutions:} Another line of work, such as in \cite{liu2019_feature,gong2019_video-face}, incorporates mechanisms to aggregate deep learning-based features of a given face track to provide a single track level embedding, which is in turn used to perform non-temporal face clustering. Recent approaches, such as \cite{huo2020_unique_faces}, adopt 3D convolutions inside the feature extractor to model temporal identity information better. Our method utilizes temporal information in a more flexible way, thus allowing the use of any feature encoder architecture.

\begin{figure}[t]
\includegraphics[scale=0.9,width=12.2cm]{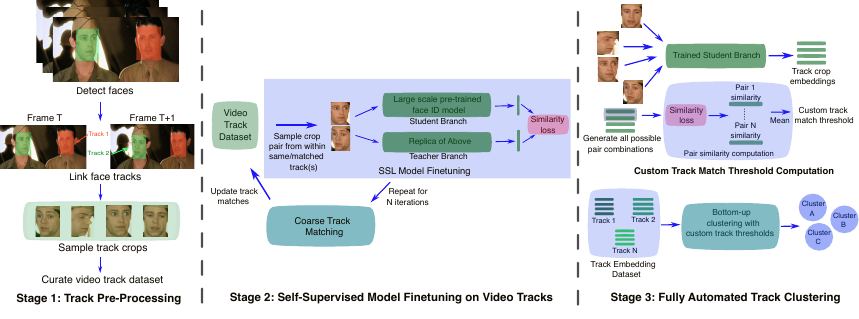}
\centering
\caption{\textbf{Overview of VideoClusterNet:} Stage 1: Given the temporal continuity in the video domain, faces detected in consecutive frames are first locally grouped into tracks using a motion tracking algorithm. Stage 2: A large-scale pre-trained face ID model is finetuned on these tracks using temporal self-supervision (w/ only positive pairing), via learning through natural and augmented face variations within each track. The finetuning is bootstrapped by soft-matching tracks across common identities. Performing these two steps alternatively helps the model better understand the given set of faces. Stage 3: An agglomerative clustering algorithm based on a model-learned similarity metric groups common identity tracks.\looseness=-1}
\label{fig:system_overview}
\end{figure}

\textbf{Temporal Track Constraints based Video Face Clustering:} A large body of methods focuses on generating identity labels through the creation of positive image pairs. They track a given face across consecutive frames and negative pairs through co-occurring tracks. Approaches such as \cite{cherniavsky2010_semi-supervised,kapoor2009_which-faces,yan2006_discriminative} apply such temporal constraints in semi/fully supervised settings, whereas methods like \cite{wu2013_simultaneous-clustering,wu2013_constrained-clustering,tapaswi2014_total-cluster,xiao2014_weighted-block,dahake2021_face-recognition,tapaswi2019_video-face, Sharma2020_clustering,aggarwal2022robust,datta2018_unsupervised-learning,kalogeiton2020_videofaceclustering,wang2023_video_centralized} use temporal constraints in an unsupervised manner, with the majority of them adopting some contrastive pair loss formulation. We significantly improve on this major trend by skipping negative pairs selection and, thus, any complex mining strategy for obtaining them.  

\textbf{Deep Metric Learning:} Deep face clustering inherently relies on having embeddings of the same identity closer to each other and of different identities farther away in the representation space. 
Approaches such as \cite{song2017_deep_metric,cinbis2011_unsupervised-metric,zhang2016_metric_learning,yang2016_joint-unsupervised,law2017_spectral,mensink2012_metric_learning} focus on optimizing such a space and require defining a face similarity metric to improve video face clustering performance.
Most approaches incorporate a contrastive-based similarity metric, such as triplet loss, to help obtain an embedding space optimized for a given set of faces in a video. We adopt the metric defined in \cref{equation:ssl_loss} that does not rely on any negative pairing, thereby avoiding any sub-optimality induced through incorrect negative pair selection. In this paper, we also utilize this metric for final video track clustering, which provides enhanced performance since the embeddings are optimized w.r.t. the metric itself. 

\textbf{Joint Representation and Clustering:} \cite{sharma2019_self-supervised} proposes two methods, TSiam and SSiam, to generate positive and negative pairs for model finetuning on a set of given video faces. TSiam adopts track-level constraints, i.e., sampling faces within the same and co-occurring tracks for positive and negative pairing, respectively. SSiam mines hard contrastive pairs using a pseudo-relevance feedback (pseudo-RF) inspired mechanism \cite{yan2006_discriminative}. Both methods employ complex modules that depend on finetuned parameters to mine negative pairs. In contrast, our proposed method does not depend on negative pairs at all, making it much simpler and more generalized. 

\cite{zhang2016_joint-face} incorporates a Markov Random Field (MRF) model to assign coarse track cluster labels, used as weak supervision for iteratively training a feature encoder. Negative pairs are mined through specific temporal constraints to boot start optimization of MRF. Unlike related prior works, we present a model adaptation stage that is highly effective due to the self-distillation procedure and its sole dependence on positive match pairs. Our final clustering algorithm directly benefits from the model finetuning stage by incorporating the self-supervised training objective as a distance metric. Such a learned metric helps boost clustering performance as it evaluates inter-track distances in an embedding space, which is explicitly optimized for reduction in observed same identity distances.


\section{Method}\label{sec:method}

\subsection{Overview}
A high-level overview of our proposed method is shown in \cref{fig:system_overview}.
The following subsections describe in detail the prominent components of our approach.

\subsection{Face Track Pre-Processing}
\label{subsec:track_preprocessing}  

 In a standard frame rate video, frame content within the same scene gradually varies w.r.t. its temporal neighboring frames.
 To exploit this temporal stability for face clustering, we first locally cluster detected faces in a video by motion tracking,
 akin to all major prior works \cite{cinbis2011_unsupervised-metric,sharma2019_self-supervised,wu2013_constrained-clustering}.
This pre-processing stage consists of four components.\looseness=-1
 
 First, scene cuts are detected in the given video, which divides it into contiguous separate sections, here coined as shots. Each scene cut represents a major change in scene composition, either involving a camera angle or scene setting change. We employ a threshold-based scene cut detection algorithm implemented in PySceneDetect library \cite{pyscenedetect}. Second, we utilize a face detection algorithm to find all visible faces in each frame of the processed shots. We adopt RetinaFace \cite{deng2020_retinaface} as our face detector due to its current state-of-the-art benchmark performance. 

 Third, the detected face crops are evaluated for their face ID quality by thresholding facial attributes based on blurriness and crop size. Crops failing the quality test are directly labeled as \textit{Unknown}. Fourth, detected faces within a given shot are locally linked into a face track using a motion tracking algorithm. We adopt the state-of-the-art method, BoT-SORT \cite{Aharon2022_bot-sort}, to generate tracks. For each face track $t$, a face crop $I$ is sampled every 12-th frame, i.e., $t = [I_{t_1},I_{t_2},....,I_{t_n}]$, where $t_n = 12*n+f_1$ and $f_1$
 denotes the original frame index for the first frame in the track's sampled set $t$.
 This particular frame interval assumes a video frame rate of $24$ fps and ensures, in most cases, that there is a significant change in either facial pose and/or expressions through the track duration.  

\subsection{Task Objective Formulation}
\label{subsec:task_formulation}
Following \cite{gong2019_video-face,wu2013_simultaneous-clustering}, we consider a set of all detected tracks within a given video, which can be denoted by $T=\{t_j | j=1,2,....,N\}$ for a set of N tracks. The face clustering objective can be defined as assigning an unique cluster id $d$ for each track $t_j$, where all tracks with the same id belong to an unique facial identity in the set $T$.
Note that the ground truth number of clusters is undefined. More formally,

\begin{equation}
    t_j^d = \{-1,1,2,...\}, \; \forall j = \{1,2,...,N\},
\end{equation}

where $d=-1$ indicates the \textit{Unknown} face cluster. This cluster represents tracks with the majority of their faces flagged as failures in any of the previously mentioned face attributes tests or if the face ID model was uncertain about it. The latter case is detailed in \cref{subsec:face_quality_estimation}.  

\subsection{Self Supervised Model Finetuning}
\label{subsec:model_finetuning}

To adapt a large-scale pre-trained face ID model to a specific set of faces, we incorporate the notion of finetuning the model for that face track set. Traditional supervised finetuning would require human supervision, i.e. ground truth labels, which can be tedious depending on the number of tracks involved.
To alleviate this problem, several approaches in the domain of self-supervised feature learning (SSL) have recently been proposed \cite{oord2018_representation_learning,tian2020_contrastive,he2020_momentum_coding,chen2020_simple}.
Especially interesting are methods that only use positive pairs for contrastive-based learning \cite{grill2020_bootstrap,zhou2022_mugs}.
Inspired by \cite{zhou2021_ibot}, we adopt a self-distillation-based SSL method that uses a teacher-student mechanism and positive pairs.\looseness=-1

\begin{wrapfigure}[17]{R}{0.59\textwidth} 
\includegraphics[scale=0.5,width=7.1cm]{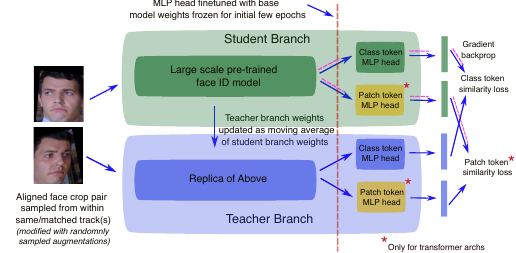}
\centering
\caption{\textbf{Self-Supervised Model Finetuning:} Face crop pair sampled from within same/matched track(s) is passed through a student and teacher branch, respectively. Gradients w.r.t. similarity loss are backpropagated only through the student branch, while the teacher weights are updated as moving average of student weights. Random augmentation set includes horizontal flipping, rotation, and color temperature variations.}
\label{fig:ssl_finetuning}
\end{wrapfigure}

First, we modify the technique to perform finetuning rather than training from scratch. As shown in \cref{fig:ssl_finetuning}, given the pre-trained face-ID model, which has no specific architecture limitations, we attach a randomly initialized multilayer perceptron (MLP) as a model head. For a Transformer model architecture \cite{Alexey2021_image}, separate heads are attached for the class and patch token embeddings, respectively. The base model with the attached head(s) is duplicated to create a teacher branch, with the original one designated as the student branch.  

Second, image pairs are curated from each track's sampled crop set. Images in a given pair are first passed through a set of augmentations, 
then an embedding pair is obtained as the output of the two constructed branches. 
We adopt a similarity loss to compare these embeddings, presented as follows:

\begin{equation}
\label{equation:ssl_loss}
    L_{ssl} = -1*softmax\left(\frac{embed_{t} - c}{temp}\right)*\log(softmax(embed_{s})),
\end{equation}

where $embed_t$ and $embed_s$ are the embeddings from teacher and student branches, respectively. Here, $c$ denotes a rolling average teacher embedding computed across training batches, and $temp$ is a fixed softening temperature. Respective loss gradients are back-propagated through the student branch weights only, while the teacher branch weights are updated via a moving average of the student weights at regular training intervals. Since the model is finetuned on all face tracks that need to be clustered in a self-supervised fashion, the training and validation sets are identical\footref{note1}. 


As the branch heads are randomly initialized, each of the branch's base model is frozen for an initial training phase. First, the heads are updated separately, akin to the description above. Then, both the base model and the heads are updated. Such a structured training regime encourages the model branch heads to produce robust and consistent embeddings for a given facial identity across the observed range of poses, expressions, lighting and appearance changes, thus improving overall clustering performance for that specific video. Model finetuning hyperparameters such as training epochs, batch size, and learning rate\footnote{\label{note1}Refer supplementary material for further details} are kept constant and are agnostic to a given dataset's attributes. 

Specific image augmentations\footref{note1} are applied during the finetuning process. This ensures that the model observes enhanced facial variations during its finetuning and helps reduce its primary dependence on natural variations present in the video. \cref{table:dataset_comparisons} provides empirical evidence in support of this fact wherein the track count ranges between 119 and 917 across our dataset movies. Our method manages to achieve significant performance improvements over benchmark methods across all movies despite these notable track count variations, evident in \cref{table:moviefacecluster_results}.

\subsection{Coarse Track Matching}
\label{subsec:track_matching}

Since a given face track is limited to being within a shot, there is no significant variation in lighting and/or face appearance across the track, which theoretically puts an upper bound on the model learning capacity.

\begin{wrapfigure}[17]{R}{0.56\textwidth} 
\includegraphics[scale=0.6,width=6.58cm]{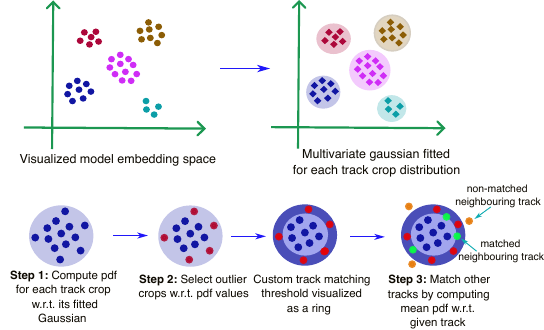}
\centering
\caption{\textbf{Coarse Face Track Matching:} A Multivariate Gaussian is fitted to every track crop distribution. Then, a custom track matching threshold is computed using outlier crops pdf values. Neighboring tracks having a mean pdf value higher than a custom threshold are soft-matched with the given track.}
\label{fig:track_matching}
\end{wrapfigure}

However, for real-world scenarios such as those likely to be observed in movies and TV series, such parameters can vary greatly throughout the video. 
To account for such variations and facilitate further model learning, we perform fully automated coarse matching of tracks across the entire dataset. 
Image pairs generated from such coarse-matched tracks enable the model to better adapt to specific lighting and appearance variations encountered in a given face across the entire dataset.
This notion is supported by our experimental findings in \cref{sec:results}.
 
For coarse track matching, 
we leverage multiple sampled crops of the same identity in a given face track. To model this track crop distribution, we found empirically that fitting a multivariate normal distribution on all track crop embeddings works the best. Mathematically, we adopt

\begin{equation}
\label{equation:mutivariate_normal}
\resizebox{.9\hsize}{!}{$
    N_{t_j}(\mu_{t_j},\Sigma_{t_j}) = (2\pi)^{-d/2}*\det(\Sigma_{t_j})^{-1/2}*\exp\left({\frac{-1}{2}(x-\mu_{t_j})^{\top}}\Sigma_{t_j}^{-1}(x-\mu_{t_j})\right),$}
\end{equation}

where $\mu_{t_j} \in \mathbb{R}^{d}$ and ${\Sigma_{t_j} \in \mathbb{R}^{d \times d}}$ are the mean and covariance matrix for the $j^{th}$ face track, computed using all its sampled face crop embeddings. Here, $d$ denotes the number of dimensions of the fitted distribution, which equals the dimension of track crop embeddings. 

To automatically set a custom threshold value for matching a given face track to other neighboring tracks in the model learned embedding space, we resort to the probability density function (pdf) values of a given track's crop embeddings. Specifically, the pdf values of all crop embeddings are computed w.r.t. their parent track's fitted distribution using \cref{equation:mutivariate_normal}. We then consider the lowest 25\% of these values and compute their mean. Experimental values ranging from 5\% to 40\% were considered, with value of 25\% empirically providing optimal true positive matches and avoiding any false positive matches. This provides a customized matching threshold, which is illustrated as a ring around the track's distribution in \cref{fig:track_matching}.
To get coarse matches for a given track, we compute a mean embedding for every other dataset track and its corresponding pdf value w.r.t. the given track's fitted distribution. If a neighboring track's pdf value is equal to or higher than that given track's custom match threshold,
then the track pair's distributions have significant overlap, hinting at a strong face identity match.

\begin{wrapfigure}[37]{R}{0.44\textwidth} 
  \begin{algorithm}[H]
    \SetCustomAlgoRuledWidth{0.44\textwidth}  
    \scriptsize
    \caption{
    Face Track Clustering}
      \textbf{Input:} \\
       \textbf{Filtered Face Tracks} $T=\{t_j |  j=1,2,...,N\}$ \\
         $\ni t_j=\{I_{t_1},I_{t_2},...I_{t_n} | t_{n}= 12*n + f_{1}\}$, \\
       \textbf{finetuned model} $\theta_{ft}$, \textbf{Similarity metric} $S$ \\
       \textbf{Stage 3.1 (Compute track crop embedding set):} \\
       \For{$t_j$ in $T$}{
            \For{$I_{t_n}$ in $t_j$}{
                $E_{t_n} \longleftarrow \theta_{ft}(I_{t_n})$
                }
            $t_{jE}=\{E_{t_1},E_{t_2},...,E_{t_n}\}$
       } 
       $T_{E} = \{t_{1E},t_{2E},...t_{NE}\}$ \\
       \textbf{Stage 3.2 (Compute custom track threshold):} \\
       \For{$t_{jE}$ in $T_E$}{
            $Sim_{t_{jE}} = \{S(E_{t_l},E_{t_m})$ $\forall (l,m) \in {}^{n}C_{2} \}$ \\
            $Thres_{t_j} \longleftarrow mean(Sim_{t_{jE}})$
       }
       $T_{thres} = \{Thres_{t_1},...,Thres_{t_N}\}$ \\
       \textbf{Stage 3.3 (Perform track clustering):} \\
        \textbf{Initialize} $i=0, C = \{C_{j} = \{t_{jE}\} \forall j=1,2,...,N$\} \\
        
        \Repeat{$N_{C_i}-N_{C_{i-1}} = 0$}{
            \For{$C_j$ in C}{
                \For{$C_k$ in C if $k \neq j$}{
                    $Sim_{jk} \longleftarrow mean(\{S(t_{jE_{a}},{t_{kE_{b}}})$ $\forall (a,b) \in {}^{n}C_{2}\})$ \\
                    \If{$Sim_{jk} < Thres_{t_j}$/$Thres_{t_k}$}{
                       $C_j \longleftarrow merge(C_j,C_k)$ \\
                    }
                }
            }
            $C = link\_merges (C)$ \\
            $N_{C_i} = cluster\_count(C)$ \\
            Repeat Stage 2 for new merged cluster set $C$\\
            $i = i + 1$
        }
    \textbf{Output:} \textbf{Clustered track IDs} $C$ 
    \label{alg:face_clustering}
  \end{algorithm}
\end{wrapfigure}

Further, given coarse matches for every dataset face track, we curate face crop pairs across these track matches for the next iteration of model finetuning. In particular, for each face crop in a given track, we randomly sample a track from a set of its coarsely matched tracks. The image pair is created by randomly sampling a crop from within the sampled track. We empirically found that this image pairing mechanism works better than other more complex strategies, such as thresholding inter track euclidean/cosine distances. If a track has no coarse matches, then we create pairs from within the same track.

\subsection{Track Face Quality Estimation}
\label{subsec:face_quality_estimation}

In complex face identification scenarios,
excluding bad quality crops/tracks becomes essential for coarse track matching and final clustering. Bad face quality of a given track often relates to model uncertainty,
which can result in sub-optimal performance for coarse matching and final clustering phases.
To automatically estimate the face quality of track crops, we adopt SER-FIQ \cite{terhorst2020_ser-fiq}, which utilizes a dropout layer to determine consistency in embeddings across multiple model iterations. For bad quality crops, the learned model that is uncertain about them would predict embeddings with high variance, thus resulting in a low-quality score.\looseness=-1

To compute the face quality score for a given track (tqs), we adapt SER-FIQ to work on a track level by obtaining scores for each track crop and averaging them. To detect bad quality tracks,
we adopt the median absolute deviation (MAD) \cite{leys2013_detecting} to detect outlier tracks based on dataset track score distribution and compute a threshold value\footref{note1}.
Low quality score tracks are filtered out from coarse track matching and final clustering modules and their final track cluster IDs are assigned as \textit{Unknown}.

\begin{figure}[t]
\includegraphics[scale=0.7,width=12.2cm]{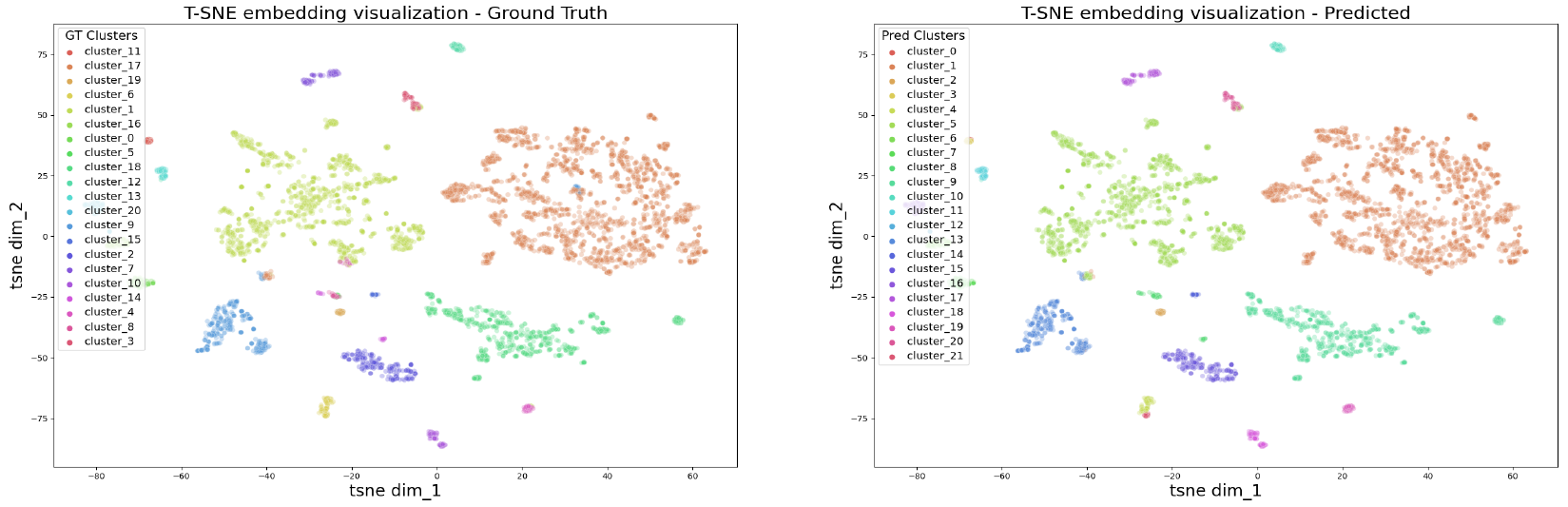}
\caption{Comparative t-SNE embedding visualizations \cite{laurens2008_tsne} on \textbf{MovieFaceCluster:The Hidden Soldier} dataset. \textit{Left:} Ground Truth (GT), \textit{Right:} Our method. Each dot in the diagram above represents the finetuned model's extracted embedding for a given face crop $I_{t_{n}}$ in a given track's sampled crop set $t$. Face embeddings assigned to a given color constitute a single cluster. Our method predicts almost perfectly the cluster designations (22 clusters) w.r.t. ground truth (21 clusters). Also note that our method correctly assigns cluster IDs to certain outlier tracks in GT. Such tracks pose a significant face clustering challenge owing to their distance from respective GT cluster centers.}
\label{fig:benchmark_viz}
\end{figure}

\subsection{Track Clustering Algorithm}
\label{subsec:final_clustering}

To cluster tracks across common identities, we utilize the SSL loss function in \cref{equation:ssl_loss} as an embedding similarity metric. Prior works incorporate Euclidean or other pre-defined distance metrics to compare model embeddings \cite{sharma2019_self-supervised,tapaswi2019_video-face}. Our proposed metric has a significant benefit. A finetuned model's embedding space is directly optimized w.r.t. this metric, thus making it optimal for evaluating embedding similarity.
As such, there are no implicit assumptions made about the space through generic distance functions. Unlike other methods that define a global matching threshold \cite{defays1977_efficient_algorithm}, 
our approach provides enhanced performance through the adaptive custom threshold computed for each track. Such a thresholding mechanism helps the algorithm automatically adjust to how effectively the model can match a given track’s identity across a true positive match pair. 


Our proposed clustering methodology is detailed in \cref{alg:face_clustering}.
We begin by creating all possible pair combinations among sampled face crops within a given track. Here, we exclude tracks filtered out by the track face quality estimation module described in \cref{subsec:face_quality_estimation}. A pair's similarity value is computed via the loss metric by passing the respective face crops through each of the model branches. As the loss metric is not commutative, a mean value is computed by alternatively sending both images through each of the branches. A custom track matching threshold is set as the average of all curated pair similarity values in the given track. This threshold represents quantitatively how well the model matches face crops belonging to a given common facial identity since the crops are part of the same track. We repeat this step for all tracks in the dataset. 

The next stage comprises merging tracks in a bottom-up approach, akin to Hierarchical Agglomerative Clustering (HAC) \cite{defays1977_efficient_algorithm}. Initially, each track is assigned an individual cluster, and tracks are iteratively merged if they satisfy a matching criterion. 
It involves creating all pairwise combinations of face crops across both tracks for a given candidate track pair.
Given the similarity (loss) value for every crop pair combination, taking a mean across them provides a matching potential value for that candidate track pair.
If this match potential is lower than either of the track's custom match thresholds, then the track pair is considered a positive match. This process is repeated for all possible track combinations in the dataset. All positively matched pairs are searched for common tracks so that they can be combined together into a bigger cluster. For example, if track pairs (1,2) and (2,3) are matched, then tracks 1, 2, and 3 are combined together. This entire matching process is run in an iterative fashion. For later iterations, where clusters could have more than one track, mean track embeddings are considered instead of a combined set of face crops for cluster pair matching,
to avoid exponentially growing match computations. The algorithm terminates when no new clusters are merged in the new iteration.
\section{Results}
\label{sec:results}
In this section, we present our experimental analysis on popular benchmark datasets and our curated movie dataset.  

\subsection{Benchmark Datasets}
Following prior work \cite{bauml2013_semi-supervised,tapaswi2019_video-face,sharma2019_self-supervised}, we evaluate our proposed method on TV series episodes of Big Bang Theory (BBT) and Buffy The Vampire Slayer (BVS), specifically the first six episodes of BBT season 1 and BVS season 5, respectively.
BBT is a TV series with a primarily indoor setting, a cast of 5$\sim$8 different characters, and 625 average face tracks per episode.
Here, all shots include wide-angle scenes, and faces are relatively small. The most common face ID challenges are pose and lighting variations.
BVS poses different challenges. The main cast comprises 12$\sim$18 characters, and there are 919 average face tracks per episode. Shots are mainly captured outdoors, and scenes are dark. It also has more close-up shots and, thus, larger face sizes.
Detailed statistics on these datasets can be found in Tab. 1 in \cite{bauml2013_semi-supervised}. 
\cref{table:BBT_BVS_results} compares our method with state-of-the-art methods on BBT and BVS, respectively.

\begin{minipage}[c]{0.45\textwidth}
\scalebox{0.59}{
\begin{tabular}
{P{2.1cm}|P{0.9cm}|P{0.9cm}|P{0.9cm}|P{0.9cm}|P{0.9cm}|P{0.9cm}|P{1.5cm}}
 \textbf{Method} & \multicolumn{7}{|c}{\textbf{BBT S01 Episode}} \\\cline{2-8}
        & S1E1 & S1E2 & S1E3 & S1E4 & S1E5 & S1E6 & Combined \\
 \hline
  SCTL \cite{wu2013_simultaneous-clustering} & 66.48 & - & - & - & - & - & - \\
  TSiam \cite{sharma2019_self-supervised} & 96.4 & - & - & - & - & - & - \\ 
  SSiam \cite{sharma2019_self-supervised} & 96.2 & - & - & - & - & - & - \\ 
  MLR \cite{bauml2013_semi-supervised} & 95.18 & 94.16 & 77.81 & 79.35 & 79.93 & 75.85 & 83.71 \\ 
  BCL \cite{tapaswi2019_video-face} & 98.63 & 98.54 & 90.61 & 86.95 & 89.12 & 81.07 & 89.63\\
  CCL \cite{Sharma2020_clustering} & 98.2 & - & - & - & - & - & - \\
  VCTRSF \cite{wang2023_video_centralized} & 99.39 & \textbf{99.84} & 97.58 & 96.41 & 98.47 & 93.33 & 94.20\\
  \hline \hline\rowcolor{LightCyan}
  Ours$\star$$\dagger$ & \textbf{99.70} & 99.67 & \textbf{98.60} & \textbf{98.80} & \textbf{99.10} & \textbf{97.10} & \textbf{98.70} \\
\end{tabular}}
\end{minipage}
\qquad
\begin{minipage}[c]{0.45\textwidth}
\scalebox{0.53}{
\begin{tabular}{P{2.2cm}|P{0.9cm}|P{0.9cm}|P{0.9cm}|P{0.9cm}|P{0.9cm}|P{0.9cm}|P{1.5cm}}
 \textbf{Method} & \multicolumn{7}{|c}{\textbf{BVS S05 Episode}} \\\cline{2-8}
        & S5E1 & S5E2 & S5E3 & S5E4 & S5E5 & S5E6 & Combined \\
 \hline
HMRF \cite{wu2013_constrained-clustering} & - & 50.3 & - & - & - & - & - \\
WBSLRR \cite{xiao2014_weighted-block} & - & 62.7 & - & - & - & - & - \\ 
TSiam \cite{sharma2019_self-supervised} & - & 92.46 & - & - & - & - & - \\ 
SSiam \cite{sharma2019_self-supervised} & - & 90.87 & - & - & - & - & - \\ 
CP-SSC \cite{somandepalli2019_reinforcing} & - & 65.2 & - & - & - & - & - \\
MvCorr \cite{somandepalli2021_robust} & - & 97.7 & - & - & - & - & - \\
MLR \cite{bauml2013_semi-supervised} & 71.99 & 61.27 & 66.60 & 67.07 & 69.59 & 61.72 & 66.37 \\ 
 BCL \cite{tapaswi2019_video-face}   & 92.08 & 79.76 & 84.00 & 84.97 & 89.05 & 80.58 & 83.62 \\
 CCL \cite{Sharma2020_clustering} & - & 92.1 & - & - & - & - & - \\\hline \hline \rowcolor{LightCyan}
  Ours$\star$$\dagger$ & \textbf{96.30} & \textbf{99.10} & \textbf{98.70} & \textbf{97.43} & \textbf{99.00} & \textbf{96.78} & \textbf{96.10} \\
\end{tabular}}
\end{minipage}
\captionof{table}{WCP/Clustering Accuracy on BBT-S01 and BVS-S05. $\star$We use ArcFace-R100 \cite{Jiankang2019_arcface} as our pre-trained base model. Combined results indicate clustering performance on set of face tracks from all six episodes combined together. $\dagger$ For fair literature comparison, we use the same face detection, tracking, and clustering labels as provided in \cite{tapaswi2019_video-face,sharma2019_self-supervised}, thereby not utilizing our proposed advanced pre-processing modules in order to effectively compare pure track clustering performance against literature methods.}
\label{table:BBT_BVS_results}

\scalebox{0.59}{
\begin{tabular}
{P{3.2cm}|P{4.5cm}|P{2.4cm}|P{2.4cm}|P{4.5cm}|P{2.7cm}}
     & & \multicolumn{4}{|c} {\textbf{Attribute}} \\\cline{3-6}
    \textbf{Dataset} & \textbf{TV series Episode/Movie} & \textbf{Unique characters} & \textbf{Track Count} & \textbf{Unique ethnicity} (other than White/Caucasian) (cast percentage) & \textbf{Avg. Track Face Quality Score $\dagger$} \\\hline\hline
        & S01E01 & 6 & 647 & A (Minor) & 0.7193 \\
        & S01E02 & 5 & 613 & A (Minor) & 0.7108\\
    The Big Bang    & S01E03 & 7 & 562 & A (Minor) & 0.7094\\
    Theory (BBT)   & S01E04 & 8 & 568 & A (Minor) & 0.7140\\
        & S01E05 & 6 & 463 & A (Minor) & 0.7177\\
        & S01E06 & 6 & 651 & A (Minor) & 0.7111 \\\cline{3-6}
        &        & Average: 6.33 & Total: 3504 & Unique Count: 1  & Average: 0.714 \\\hline\hline
        & S05E01 & 12 & 786 & None & 0.7090 \\
        & S05E02 & 13 & 866 & None & 0.7117\\
Buffy The & S05E03 & 14 & 1185 & None & 0.7150\\
Vampire Slayer & S05E04 & 15 & 852 & None & 0.7125\\
  (BVS) & S05E05 & 15 & 733 & None & 0.7081\\
        & S05E06 & 18 & 1055 & None & 0.7142 \\\cline{3-6}
        &        & Average: 14.5 & Total: \textbf{5477} & Unique Count: 0 & Average: 0.712 \\\hline\hline
        & An Elephant's Journey & 18 & 562 & None & 0.7112 \\
        & Armed Response & 14 & 119 & AA (Major), ME (Major) & 0.7085 \\
        & Angel Of The Skies & 29 & 319 & None & 0.7150\\
       & Death Do Us Part (2019) & 7 & 395 & AA (Major) & 0.7177 \\
 MovieFaceCluster        & American Fright Fest & 37 & 457 & AA (Minor) & 0.7098\\
        & The Fortress & 64 & 917 & A (Major) & 0.6918\\
        & Under The Shadow & 9 & 143 & ME (Major) & 0.7134\\
        & The Hidden Soldier & 21 & 594 & A (Major) & 0.7056\\
        & S.M.A.R.T. Chase & 10 & 113 & A (Major) & 0.7110\\\cline{3-6}
        &        & Average: \textbf{23.2} & Total: 3619 $\dagger\dagger$ & Unique Count: \textbf{3} & Average: \textbf{0.706}\\\hline
\end{tabular}}
\captionof{table}{Specific dataset attribute comparisons across BBT, BVS and our MovieFaceCluster dataset. A: Asian, AA: African American, ME: Middle Eastern characters. $\dagger$ Score is computed as the average of all track quality scores as part of a given TV series episode/movie. A given track quality score is computed as the average of face quality scores for each of its sampled crops, using SER-FIQ \cite{terhorst2020_ser-fiq} and  ArcFace-R100 \cite{Jiankang2019_arcface} as the pre-trained model for extracting embeddings. Our lower average face quality score indicates that MovieFaceCluster contains on average more challenging cases for face clustering compared to literature datasets. $\dagger\dagger$ Only tracks containing decent quality face crops were added as part of each movie dataset, resulting in a slightly lower track count. Please refer to the supplementary material for further dataset analysis and comparisons.}
\label{table:dataset_comparisons}

\subsection{Metrics}
We define two primary metrics at face track level for evaluating video face clustering performance, namely Weighted Cluster Purity/Accuracy (WCP) and Predicted Cluster Ratio (PCR). WCP is defined as the fraction of common identity tracks in a predicted cluster, weighted by the cluster track count. PCR is the ratio between the predicted cluster and ground truth cluster count. Note that a ratio closer to 1 is deemed better. 

\subsection{Movie Dataset}
Mainstream movies present challenges for face clustering due to extreme pose, illumination, and appearance variations. Considering the lack of significant benchmark datasets in the academic research community, we present a new movie benchmark dataset named \textbf{MovieFaceCluster} \footnote{\label{note3} The MovieFaceCluster dataset can be downloaded from \href{https://www.flawlessai.com/dataset/}{here}}, 
containing a collection of movies, hand-selected by film post-production specialists, with unique face clustering challenges  \footref{note1}. Given the set of movies, we run the preprocessing mentioned in \cref{subsec:track_preprocessing} to obtain a specific track dataset for each movie. We hand-label each track with an ID using the main character cast from that track's parent movie. Also, false detections and extreme unidentifiable tracks are pre-filtered out using the track face quality estimation module mentioned in \cref{subsec:face_quality_estimation}. \cref{table:dataset_comparisons} provides detailed statistics and comparisons of this curated dataset versus literature datasets. Overall, our presented dataset provides more unique characters, comparable track count, higher ethnic diversity and more challenging face tracks compared to publicly available datasets.

\cref{table:moviefacecluster_results} also compares our method to other state-of-the-art (SoTA) approaches on this benchmark to provide empirical evidence of our algorithm's effectiveness. We implemented all five literature methods due to the lack of their public code implementations \footref{note1}. \cref{fig:benchmark_viz} shows a t-SNE visualization of our learned embeddings and clustering performance on one dataset movie, and \cref{fig:benchmark_hard_case_viz} illustrates some hard case clustered tracks using our proposed method on MovieFaceCluster.

\scalebox{0.54}{
\begin{tabular}
{P{2.1cm}|P{2.15cm}|P{2.15cm}|P{2.15cm}|P{2.15cm}|P{2.15cm}|P{2.15cm}|P{2.15cm}|P{2.1cm}|P{2.1cm}}
 & \multicolumn{9}{|c}{\textbf{Movie}} \\\cline{2-10}
  &  An &  Armed & Angel & Death & American & The & Under & The & S.M.A.R.T. \\
    \textbf{Method} & Elephant's & Response & Of The &  Do Us &  Fright &  Fortress &  The & Hidden & Chase \\
           & Journey (2019) & & Skies & Part (2019) & Fest &  & Shadow & Soldier \\\cline{2-10}
  & \multicolumn{9}{|c}{\textbf{Weighted Cluster Accuracy (\%) \& Pred Cluster Ratio (Pred / GT)} } \\\hline 
 TSiam \cite{sharma2019_self-supervised} & 90.7 \& 1.44 & 84.9 \& 1.36 & 77.1 \& 0.62 & 92.9 \& 1.57 & 89.3 \& 0.83 & 68.6 \& 0.69 & 71.8 \& 2.11 & 90.7 \& 1.33 & 79.6 \& 1.70 \\
 SSiam \cite{sharma2019_self-supervised} & 88.1 \& 1.61 & 86.6 \& 1.21 & 75.5 \& 0.59 & 94.4 \& 1.28 & 86.2 \& 0.78 & 71.1 \& 0.73 & 68.3 \& 2.33 & 88.7 \& 1.24 & 82.3 \& 1.80 \\
 JFRAC \cite{zhang2016_joint-face} & 91.4 \& 1.33 & 85.2 \& 1.50 & 73.4 \& 0.62 & 90.8 \& 0.71 & 91.5 \& 0.86 & 65.3 \& 0.77 & 73.1 \& 2.00 & 92.6 \& 1.19 & 85.8  \& 1.70 \\
 CCL \cite{Sharma2020_clustering} & 89.5 \& N.A.$\dagger$ & 89.7 \& N.A.$\dagger$ & 75.0 \& N.A.$\dagger$ & 95.4 \& N.A.$\dagger$ & 87.2 \& N.A.$\dagger$ & 62.7 \& N.A.$\dagger$ & 77.4 \& N.A.$\dagger$ & 84.0 \& N.A.$\dagger$ & 89.9 \& N.A.$\dagger$ \\
 VCTRSF \cite{wang2023_video_centralized} & 96.3 \& N.A.$\dagger$ & 92.2 \& N.A.$\dagger$ & 77.7 \& N.A.$\dagger$ & 96.5 \& N.A.$\dagger$ & 91.3 \& N.A.$\dagger$ & 78.8 \& N.A.$\dagger$ & 78.7 \& N.A.$\dagger$ & 94.4 \& N.A.$\dagger$ & 88.4 \& N.A.$\dagger$ \\\hline\hline\rowcolor{LightCyan}
 Ours & \textbf{97.2 \& 1.11} & \textbf{94.1 \& 0.93} & \textbf{85.9 \& 0.72} & \textbf{98.0 \& 1.14} & \textbf{97.6 \& 0.92} & \textbf{89.3 \& 1.02} & \textbf{82.5 \& 1.88} & \textbf{98.5 \& 1.04} & \textbf{93.8 \& 1.50} \\\hline
\end{tabular}}
\captionof{table}{Quantitative comparisons on each MovieFaceCluster dataset movie. For a fair comparison, we incorporate ArcFace-R100 \cite{Jiankang2019_arcface} as the pre-trained feature extractor for all reported methods, including ours. We outperform SoTA methods w.r.t. cluster accuracy and predicted cluster ratio. Details on our implementation of all comparative methods can be found in the supplementary material. $\dagger$Number of ground truth clusters is required as input for these methods, so PCR isn't a valid performance metric while also being a major limitation for these methods.}
\label{table:moviefacecluster_results}
\section{Ablative Analysis}\label{sec:ablation}
We ablate the central components of our method and analyze limitations and future directions.

\paragraph{Model Finetuning}
\label{para:ablation_iterative_finetuning}
We ablate on the effectiveness of generic face ID model finetuning to a given set of face tracks as part of our proposed method. \cref{table:ablation_iterative_finetuning} provides a comparison of clustering performance w/ and w/o using the model finetuning module. Note that our proposed clustering algorithm depends on the similarity metric learned during the finetuning stage. As such, to compare both methods in a fair way, we adopt a baseline clustering algorithm, i.e., HAC with average linkage and cosine distance metric. Performing model finetuning results in roughly 6\% increase in cluster accuracy, which underlines its usefulness.  

\paragraph{Final Clustering Algorithm}
\label{para:ablation_final_clustering}
We further ablate on the performance of our final clustering approach vs. baseline algorithm, i.e., HAC in \cref{table:ablation_clustering_algorithm}. Here, we keep the model finetuning stage constant in all methods to compare fairly. As for HAC,
we take the mean of a given track's sampled crop embeddings to obtain a representative track embedding.
We further ablate on the loss function as a similarity metric. Specifically, we compare our final clustering algorithm with one that uses Euclidean and Cosine distances as similarity metrics. Our approach involving the loss metric outperforms all other methods.

\paragraph{Generic face ID Model Architectures}
\label{subsec:ablation_model_architecture}
In \cref{table:ablation_face_id_model}, we ablate on our approach's generalization capabilities to incorporate any generic face ID model, fairly agnostic to its architecture class. Specifically, we compare some prominent face ID models from both CNN and Transformer architecture classes incorporated as part of our method, against using them (w/o finetuning) along with the baseline clustering method (HAC).
Regardless of the incorporated face ID model, our finetuning method provides roughly a $5\sim12\%$ performance boost, which underlines our method's capability to adapt to and improve any generic face ID model.  

\paragraph{Limitations and Future Work} 
The use of a generic face ID model means that any pre-existing model biases may also be propagated through our method. For example, if the generic model has learned an incorrect similarity between two distinct facial identities, then our algorithm might adapt to it and provide a false positive cluster for that given pair. A future direction could be to automatically detect such biases, such that a given pair's embeddings are specifically pulled apart. This could be done by incorporating an outlier detection technique based on pair similarity values for a cluster's tracks. Also, given that we finetune on a set of face tracks, it might not be optimal for real-time applications depending on the track set size. 

\begin{minipage}[c]{0.25\textwidth}
\scalebox{0.52}{
\begin{tabular}{p{4.1cm}|P{1.4cm}}
 \textbf{Method} & \textbf{Cluster} \\ 
        & \textbf{Accuracy (\%)} \\
 \hline
 Baseline-(Non-Finetuned) & 86.10 \\\rowcolor{LightCyan}
  \textbf{Ours-(Finetuned)} & \textbf{91.52} \\\hline
\end{tabular}}
\captionof{table}{Ablation for model finetuning module, using ArcFace-R100 as the base model. Experiments are performed on \textbf{MovieFaceCluster: The Hidden Soldier} dataset and HAC as the final clustering algorithm.}
\label{table:ablation_iterative_finetuning}
\end{minipage}
\quad
\begin{minipage}[c]{0.34\textwidth}
\scalebox{0.51}{
\begin{tabular}{P{1.9cm}|P{1.9cm}|P{1.7cm}|P{2.2cm}}
 \textbf{Clustering} & \textbf{Similarity} & \textbf{Cluster} & \textbf{Cluster}\\ 
\textbf{Algorithm} & \textbf{Metric} & \textbf{Accuracy (\%)} & \textbf{Ratio (Pred/GT)}\\
 \hline
 Baseline (HAC) & Cosine & 91.52 & 1.43 (30/21) \\
 Ours & Cosine & 93.70 & 2.0 (42/21) \\
 Ours & Euclidean & 96.50 & 3.5 (74/21) \\\rowcolor{LightCyan}
 \textbf{Ours} & \textbf{Loss Func.} & \textbf{98.50} & \textbf{1.04 (22/21)}  \\\hline
\end{tabular}}
\captionof{table}{Ablation for final clustering algorithm, compared with baseline HAC, and using pre-defined metrics within our algorithm. Experiments are performed on \textbf{MovieFaceCluster: The Hidden Soldier} dataset and using ArcFace-R100 as the base model.}
\label{table:ablation_clustering_algorithm}
\end{minipage}
\quad
\begin{minipage}[c]{0.31\textwidth}
\scalebox{0.56}{
\begin{tabular}{p{3.7cm}|p{1.5cm}|>{\columncolor{LightCyan}}p{1.0cm}}
 \textbf{Face ID Model} & \multicolumn{2}{|c}{\textbf{Cluster Acc. (\%)}} \\\cline{2-3}
               & \textbf{Baseline} & \textbf{Ours} \\
 \hline
 FaRL-P16 \cite{yinglin2022_farl}        & 78.7 & \textbf{90.2} \\
 VGGFace2-R50 \cite{qiong2018_vggface2}    & 84.2 & \textbf{95.7} \\
 ArcFace-R100 \cite{Jiankang2019_arcface}    & 86.1 & \textbf{98.5} \\
 AdaFace-R100 \cite{minchul2022_adaface}    & 86.9 & \textbf{98.4} \\\hline
\end{tabular}}
\captionof{table}{Ablation for various base face ID models incorporated in our method. We perform comparisons using our proposed method and pre-trained model + baseline clustering (HAC). Experiments are performed on \textbf{MovieFaceCluster: The Hidden Soldier} dataset.}
\label{table:ablation_face_id_model}
\end{minipage}

\section{Conclusion}\label{sec:conclusion}

We present a novel video face clustering algorithm that specifically adapts to a given set of face tracks through a fully self-supervised mechanism. This helps the model better understand and adapt to all observed variations for a given facial identity across the entire video without any human-in-the-loop label guidance. Our fully automated approach to video face clustering specifically helps avoid any sub-optimal solutions that may be induced from non-intuitive user-defined parameters. In addition, using a model-learned similarity metric over generic distance functions helps provide SoTA video face clustering performance over other competing methods. Extensive experiments and ablation studies on our presented comprehensive movie dataset and traditional benchmarks underline our method's effectiveness under extremely challenging real-world scenarios.

\bibliographystyle{splncs04}
\bibliography{00.main}

\end{document}